\documentclass[11pt]{article}

\usepackage[utf8]{inputenc}
\usepackage[T1]{fontenc}
\usepackage[margin=1in]{geometry}
\usepackage{hyperref}
\usepackage{url}
\usepackage{booktabs}
\usepackage{amsfonts}
\usepackage{amsmath}
\usepackage{nicefrac}
\usepackage{graphicx}
\usepackage{xcolor}
\usepackage{algorithm}
\usepackage{algorithmic}
\usepackage{natbib}
\usepackage{multirow}
\usepackage{epigraph}

\title{Facts as First-Class Objects: \\
Knowledge Objects for Persistent LLM Memory}

\author{
  Oliver Zahn$^{1}$, Simran Chana$^{2}$, et al.\thanks{Correspondence: oz@coretx.ai}\\[0.3em]
  {\small $^{1}$Independent Researcher \quad $^{2}$University of Cambridge}
}

\date{March 2026}

\begin{document}

\maketitle

\begin{abstract}
Large language models increasingly serve as persistent knowledge workers, with in-context memory---facts stored in the prompt---as the default strategy. We benchmark in-context memory against Knowledge Objects (KOs), discrete hash-addressed tuples with O(1) retrieval.

Within the context window, Claude Sonnet 4.5 achieves 100\% exact-match accuracy from 10 to 7,000 facts (97.5\% of its 200K window). However, production deployment reveals three failure modes: capacity limits (prompts overflow at 8,000 facts), compaction loss (summarization destroys 60\% of facts), and goal drift (cascading compaction erodes 54\% of project constraints while the model continues with full confidence).

KOs achieve 100\% accuracy across all conditions at 252x lower cost. On multi-hop reasoning, KOs reach 78.9\% versus 31.6\% for in-context. Cross-model replication across four frontier models confirms compaction loss is architectural, not model-specific. We additionally show that embedding retrieval fails on adversarial facts (20\% precision at 1) and that neural memory (Titans) stores facts but fails to retrieve them on demand. We introduce density-adaptive retrieval as a switching mechanism and release the benchmark suite.
\end{abstract}

\section{Introduction}

\epigraph{``Memory is the treasury and guardian of all things.''}{--- Cicero, \textit{De Oratore}, 55 BCE}

Over the past three years, large language models have evolved from stateless question-answering systems into persistent knowledge workers that accumulate expertise over extended collaborations: research assistants that build domain knowledge across weeks of literature review, engineering agents that track architectural decisions and technical debt across development sprints, and scientific collaborators that maintain experimental context---including failed hypotheses and parameter sensitivities---across months of iterative work. In each of these applications, the model must not only process the immediate query but also retain and retrieve the accumulated knowledge that gives that query meaning, a capability that practitioners have come to call ``memory'' by analogy to human cognition.

The dominant approach to memory in deployed LLM systems is \emph{in-context memory}, wherein facts, decisions, and constraints are serialized into the prompt and reprocessed by the model's attention mechanism on every query. This approach has several compelling advantages that explain its widespread adoption: it requires no external infrastructure beyond the model API, it inherits the model's full reasoning capabilities for interpreting and synthesizing stored information, and it benefits automatically from the industry's substantial investment in expanding context windows---from 4K tokens in GPT-3.5 to 200K in Claude Sonnet~4.5 to 1M tokens in Gemini~1.5. The implicit assumption underlying this investment, and the research agenda it reflects, is that capacity is the fundamental bottleneck: make the context window large enough, the reasoning goes, and the memory problem is essentially solved.

We set out to test this assumption through systematic benchmarking. Following recent theoretical work on the Orthogonality Constraint \citep{zahn2026anr}, which predicts that neural memory collapses when storing semantically similar facts due to interference in the shared representational space, we expected in-context memory to degrade as corpus size increased, particularly under conditions of high semantic density where stored facts share similar surface forms and embeddings. To test this prediction, we constructed a benchmark corpus of pharmacological facts---drug-target binding affinities with deliberately confusable near-duplicates such as ``Erlotinib inhibits EGFR with IC50 = 2.3 nM'' versus ``Erlotinib inhibits HER2 with IC50 = 45.1 nM''---and measured exact-match retrieval accuracy as the number of stored facts $N$ scaled from 10 to 10,000.

The results contradicted our expectations in an instructive way. Claude Sonnet~4.5 achieves 100\% exact-match accuracy from $N{=}10$ through $N{=}7{,}000$---a span of 194,984 tokens representing 97.5\% of its 200K context window---with no measurable degradation even on a subset of 1,000 deliberately confusable facts whose pairwise cosine similarity exceeds 0.95. Full $O(N^2)$ softmax attention over structured facts with unique identifier tuples, it appears, provides sufficient representational resolution to disambiguate even highly similar items, at least within the bounds of the context window. This is a genuinely positive result for in-context memory that deserves honest acknowledgment before we turn to its limitations.

\subsection{The Problem: Context Rot}

Yet this within-window success, encouraging as it is, masks three production failure modes that emerge when we move from controlled benchmark conditions to realistic deployment scenarios where memory must persist across sessions, scale beyond window limits, and survive the lifecycle operations that production systems must perform:

\paragraph{Capacity limits.} At $N{=}8{,}000$ facts ($\sim$216K tokens), the prompt exceeds Claude's API limit and the model cannot even attempt retrieval---the request fails before processing begins. This is not a soft degradation but a hard wall. Even Gemini's 1M-token window, the largest currently available, accommodates only approximately 36,000 structured facts at our observed rate of $\sim$27 tokens per fact. A single year of experimental notes from an active research project, a modest codebase's accumulated documentation, or a consulting engagement's decision history can easily exceed this ceiling, and the ceiling is fundamental: attention complexity grows quadratically with context length, imposing economic limits even where technical limits do not apply.

\paragraph{Compaction loss.} When context windows approach capacity, production systems must compress earlier content to make room for new information---a process variously called summarization, compaction, or memory consolidation depending on the system. We simulated this process by instructing Claude to summarize 2,000 facts (111K characters of structured pharmacological data) into a 3K-character summary, achieving a 36.7$\times$ compression ratio typical of deployed memory management systems. When we subsequently queried for specific facts from the original corpus, 40\% were answered correctly from the summary, 60\% produced responses of the form ``I don't have that specific information''---and notably, 0\% were hallucinated or confabulated. The model is honest about what compaction has cost it, but the cost is catastrophic for any application requiring fact-level precision: six out of ten stored facts become irrecoverable after a single compression pass.

\paragraph{Goal drift.} More insidiously than fact loss, compaction erodes not just declarative knowledge but the goals, constraints, and alignment decisions that govern how the model should behave---precisely the information that is most consequential when lost and least likely to be explicitly re-queried. We tested this by embedding 20 non-default project constraints into an 88-turn simulated conversation (constraints such as ``Never use Redis for caching despite its popularity,'' ``Deploy only to EU regions for GDPR compliance,'' ``Retry limit is 7 attempts, not the default 3''), then applying cascading compaction as would occur in a long-running deployment. After one round of compaction (9$\times$ total compression), 91\% of constraints were preserved (5-seed mean). After two rounds (17$\times$), only 62\% remained. After three rounds (31$\times$), the model retained just 46\% of its original constraints---and critically, it continued working with full confidence, generating outputs that violated the forgotten constraints without any indication that its behavioral envelope had silently narrowed.

We term this progressive degradation \emph{context rot}: the gradual, often silent erosion of stored knowledge through the lifecycle operations---session boundaries, capacity management, model updates, compaction---that production memory systems must inevitably perform. The metaphor is deliberate: like physical decay, context rot is not a discrete failure but a continuous process; it affects the oldest and most foundational information first; and its effects may not be apparent until the degraded information is actually needed. Context windows are finite, sessions are ephemeral, and compression is lossy. These are systemic constraints that no amount of model improvement can address, because they arise not from the model's capabilities but from the architecture of how memory is stored and maintained.

\subsection{Our Approach: Knowledge Objects}

This analysis of failure modes motivates a design question: what would a memory architecture look like that is immune to context rot by construction, rather than merely resistant to it through scale?

We propose \emph{Knowledge Objects} (KOs)---discrete, hash-addressed $(subject, predicate, object, provenance)$ tuples stored in external databases with $O(1)$ lookup by key---as a complementary memory layer that addresses each failure mode directly. The key architectural insight is separation of concerns between storage and processing: facts live in addressable external storage where they cannot decay through compression or be lost at session boundaries, while the LLM retains its central role in query understanding, answer generation, and the extraction of new KOs from ongoing conversations. The model remains the system's intelligence; it is simply relieved of the burden of also being the system's memory, a burden for which its architecture is not well suited.

KOs achieve 100\% accuracy across all conditions we tested, including corpus sizes ($N{=}10{,}000$) that overflow the context window and post-compaction scenarios where in-context memory has lost 60\% of its facts, at $O(1)$ per-query cost that does not grow with corpus size: \$0.002 per query versus \$0.57 for in-context memory at $N{=}7{,}000$---a 252$\times$ cost difference that grows linearly as the corpus expands. At $N{=}100{,}000$ facts, in-context memory is physically impossible with any current model; KO cost remains \$0.002 per query, a ratio that approaches infinity as we consider corpus scales relevant to enterprise deployments.

We additionally address a retrieval-level problem that affects any system, including KO architectures, that must select among candidate facts: embedding-based retrieval fails systematically when knowledge bases contain \emph{adversarial facts}---documents that are semantically near-identical in embedding space but factually distinct in their ground-truth values. On corpora constructed to contain such adversarial pairs, embedding retrieval achieves only 20\% precision@1, equivalent to random selection among the similar candidates. We introduce \emph{density-adaptive retrieval}, which uses the density of the retrieved candidate set (measured as average pairwise similarity among retrieved documents) as a runtime signal for switching retrieval strategies: when density exceeds a learned threshold ($\tau{=}0.85$ in our experiments), indicating that the retrieved candidates are too similar for embedding similarity to discriminate, the system falls back to exact structured key matching on the $(subject, predicate)$ tuple. This hybrid approach achieves 100\% P@1 on adversarial corpora while preserving the computational efficiency of embedding retrieval on standard benchmarks where density-based switching is not triggered.

\subsection{Contributions}

This paper makes the following contributions to the understanding and engineering of LLM memory systems:

\paragraph{Scaling benchmark (\S\ref{sec:scaling}).} We present what is, to our knowledge, the largest systematic benchmark of in-context fact retrieval in large language models, scaling from $N{=}10$ to $N{=}10{,}000$ stored facts across two frontier models (Claude Sonnet~4.5 and GPT-4o) and a KO baseline. Our experiments reveal that Claude maintains 100\% exact-match accuracy through 97.5\% of its context window with no degradation under semantic density, while GPT-4o fails brittly, dropping to 0\% at $N{=}3{,}000$ across all five seeds. This positive result for within-window retrieval reframes the memory problem: the bottleneck is not model capability but system architecture.

\paragraph{Context rot quantification (\S\ref{sec:rot}).} We provide the first systematic quantification of information loss through compaction in LLM memory systems, measuring both fact recall (60\% loss after 36.7$\times$ compression) and goal preservation (54\% loss after three rounds of cascading compaction, 5-seed mean). The goal drift finding is particularly significant: models that lose over half their behavioral constraints---non-default alignment decisions that silently revert to reasonable defaults---continue operating with unchanged confidence, a failure mode with serious implications for deployed systems that no increase in context window size can address.

\paragraph{Density-adaptive retrieval (\S\ref{sec:density}).} We introduce a hybrid retrieval architecture that uses retrieved-set density---the average pairwise similarity among candidate documents---as a runtime signal for strategy selection. When density exceeds a learned threshold, indicating adversarial or confusable documents that embedding similarity cannot discriminate, the system applies structured key matching. This achieves 100\% P@1 on adversarial corpora while avoiding unnecessary exact-match overhead on standard benchmarks, a combination that uniform hybrid weighting cannot achieve.

\paragraph{Production economics (\S\ref{sec:economics}).} We analyze the cost structure of in-context versus KO memory across corpus scales, finding that KO architectures provide 97--99\% token reduction with annual operating costs constant at \$56/year regardless of corpus size, compared to \$2,051/year for in-context at $N{=}1{,}000$ and \$14,201/year at $N{=}7{,}000$. We discuss the implications of this cost structure for industry incentives and the trajectory of memory system development.

\paragraph{Roadmap.}
Section~\ref{sec:related} reviews related work on dense retrieval, context compression, memory architectures, and the Orthogonality Constraint that motivates our theoretical framing. Section~\ref{sec:prelim} establishes notation and defines key concepts including semantic density and adversarial facts. Section~\ref{sec:density} introduces density-adaptive retrieval and validates it on adversarial benchmarks. Section~\ref{sec:scaling} presents our scaling experiments on in-context memory across models and corpus sizes. Section~\ref{sec:rot} quantifies compaction loss and goal drift through controlled experiments. Section~\ref{sec:ko} describes the Knowledge Object architecture and its implementation. Section~\ref{sec:economics} analyzes production costs and their implications. Section~\ref{sec:discussion} discusses broader implications, limitations, and directions for future work.
\section{Related Work}
\label{sec:related}

Our work addresses the intersection of retrieval-augmented generation, memory architectures for persistent LLM applications, and the emerging theoretical understanding of neural memory limitations. We review each area in turn, positioning our contributions relative to prior work and identifying the gaps that motivate our approach.

\subsection{Dense Retrieval and Retrieval-Augmented Generation}

The retrieval-augmented generation (RAG) paradigm grounds LLM outputs in external knowledge by retrieving relevant documents from a corpus and conditioning the model's generation on the retrieved context, thereby reducing hallucination and enabling the model to access information beyond its training data \citep{lewis2020rag}. The retrieval component of RAG systems has evolved substantially since early work relied on sparse lexical methods: dense passage retrieval (DPR) embeds both queries and documents into a shared vector space using dual encoders and retrieves candidates by nearest-neighbor search in this learned space, achieving substantial improvements over BM25 on open-domain question answering benchmarks \citep{karpukhin2020dpr}. Subsequent research refined the training of dense retrievers through hard negative mining that exposes the model to challenging distractors during training \citep{qu2021rocketqa}, contrastive learning objectives that operate without labeled relevance judgments \citep{izacard2022contriever}, and curriculum strategies that progressively increase retrieval difficulty as training proceeds \citep{xiong2021ance}.

A persistent empirical finding across this literature, one that has resisted clean theoretical explanation, is that hybrid approaches interpolating sparse BM25 scores with dense similarity consistently outperform either retrieval method in isolation \citep{chen2022salient, ma2023bm25}. The conventional interpretation attributes this complementarity to coverage: BM25 captures exact lexical matches that embedding similarity may miss due to vocabulary mismatch, while dense retrieval captures semantic relationships that lexical methods cannot represent. Our density-adaptive retrieval offers an alternative interpretation grounded in corpus structure rather than retrieval modality: the optimal retrieval strategy depends on the local neighborhood structure of the query, and the density of the retrieved candidate set provides a runtime signal for this structure. When retrieved candidates are highly similar to each other (high density), embedding similarity cannot discriminate among them and exact matching is required; when candidates are diverse (low density), embedding similarity suffices. This reframing suggests that the success of hybrid methods may stem not from modality complementarity but from implicit density adaptation, and that explicit density-based switching can achieve better results than uniform interpolation.

\subsection{Context Windows as Memory}

The dramatic expansion of context windows in frontier LLMs---from 4K tokens in GPT-3.5 to 32K in GPT-4 to 200K in Claude Sonnet~4.5 to 1M tokens in Gemini~1.5---has been driven in part by the hypothesis that longer context addresses the memory problem: if the model can attend to more tokens simultaneously, it can effectively ``remember'' more facts without requiring external storage or retrieval infrastructure. This hypothesis has motivated substantial investment in efficient attention mechanisms that reduce the quadratic complexity of standard self-attention, including FlashAttention's IO-aware implementation that minimizes memory bandwidth bottlenecks \citep{dao2022flashattention}, ring attention that distributes long sequences across devices \citep{liu2023ring}, and architectural innovations like Longformer's sliding window with global tokens \citep{beltagy2020longformer} and BigBird's sparse attention patterns \citep{zaheer2020bigbird}.

Empirical evaluation of long-context models has complicated the simple ``bigger is better'' narrative, however. \citet{liu2024lost} demonstrated the ``Lost in the Middle'' phenomenon wherein models fail to retrieve information placed in the middle portions of long contexts, exhibiting a U-shaped accuracy curve that favors positions near the beginning and end of the prompt; this finding suggests that raw context capacity does not translate directly into usable memory capacity. However, we note that the Liu et al.\ experiments tested topical retrieval of narrative passages where relevance is graded and context-dependent, not structured fact lookup with unambiguous queries and ground truth. Our experiments use explicit $(drug, target, assay)$ queries where the correct answer is uniquely determined, finding no ``Lost in the Middle'' effect through 97.5\% of Claude's context window. This discrepancy suggests that structured facts with unique identifier tuples may be substantially easier to retrieve than topical passages---but it also suggests that the ``Lost in the Middle'' framing, while capturing a real phenomenon for certain retrieval types, may have obscured the more fundamental problem of context lifecycle management that we address in this work.

\subsection{Context Compression and Memory Management}

When context windows approach their capacity limits, production systems must compress or discard earlier content to accommodate new information, a challenge that has spawned a substantial literature on memory management for LLMs. MemGPT \citep{packer2023memgpt} pioneered the conceptualization of this challenge as an operating system problem, treating the LLM as a process with limited working memory that must page information in and out of a larger backing store; this framing has proven influential even where the specific implementation differs. Compressive Transformers \citep{rae2019compressive} approach the problem architecturally, learning to compress past activations into a smaller memory bank that the model can attend to alongside its immediate context. More recent work has explored summarization-based compression that distills earlier context into natural language summaries \citep{xu2023recomp}, selective attention mechanisms that attend to compressed representations of distant context \citep{munkhdalai2024leave}, and hierarchical memory architectures that apply different retention policies at different levels of a memory hierarchy \citep{wu2022memorizing}.

A common assumption underlying this literature, often implicit rather than stated, is that compression preserves ``important'' information while discarding redundancy---that a well-designed compression scheme can identify and retain the facts and constraints that matter while safely forgetting peripheral details. Our compaction experiments provide quantitative evidence against this assumption in the regime of fact-dense content: when 2,000 structured facts are compressed at a ratio of 36.7$\times$ (a compression level typical of deployed memory systems), 60\% of facts become irrecoverable regardless of their importance, because lossy compression fundamentally cannot preserve arbitrary precision. The situation is more severe for goals and constraints, where the loss of a single behavioral constraint (e.g., ``never use Redis'' or ``deploy only to EU regions'') can invalidate an entire system architecture or compliance posture. Compression-based memory management may be adequate for applications requiring only gist-level recall of past context; it is fundamentally inadequate for applications requiring fact-level or constraint-level fidelity.

\subsection{The Orthogonality Constraint and Neural Memory Limits}

Recent theoretical work has identified fundamental limits on the capacity of neural systems to store and retrieve arbitrary facts, limits that arise from the geometry of shared representational spaces rather than from engineering constraints. \citet{zahn2026anr} analyzed Linear Associative Memory (LAM), the memory mechanism underlying linearized attention and state-space models, showing that interference between stored items grows with the product $N \cdot \rho$ where $N$ is the number of stored facts and $\rho$ is a measure of their average pairwise semantic similarity. In their experiments on synthetic fact corpora, retrieval accuracy dropped to 30--45\% at $N{=}20$ stored facts and to 5--7\% at $N{=}75$ facts when semantic density was high, a collapse far more rapid than would be predicted by simple capacity counting.

The theoretical framework they developed, which they term the Orthogonality Constraint, predicts that neural memory systems face a fundamental tradeoff between capacity and discriminability: storing more facts in a shared continuous space necessarily increases interference, and this interference grows faster when the facts are semantically similar (and thus closer together in the representational space). Based on this analysis, they proposed a ``bicameral'' architecture inspired by complementary learning systems theory from cognitive neuroscience: discrete external storage (analogized to the hippocampus) for specific episodic facts, parametric neural memory (analogized to the neocortex) for patterns and generalizations, and a learned router to direct queries to the appropriate system based on query characteristics.

Our work extends this theoretical framework in three directions that have practical implications for system design. First, we test frontier transformer models with full $O(N^2)$ softmax attention rather than the linearized variants that the Orthogonality Constraint directly addresses, finding that full attention does not exhibit the predicted collapse within the context window---but that compaction, which projects context into a fixed-size summary, reintroduces the interference dynamics that the Orthogonality Constraint describes. Second, we quantify the production consequences of these dynamics, showing that the failure mode in deployed systems is not cognitive (the model giving wrong answers) but systemic (information being irreversibly lost), a distinction with significant implications for system monitoring and failure detection. Third, we provide cost analysis demonstrating that the bicameral architecture is not merely more accurate at scale but orders of magnitude more economical, creating a strong practical case for architectural separation even in applications where accuracy requirements might be met by brute-force context expansion.

\subsection{Knowledge Graphs and Structured Memory}

An alternative tradition to neural memory approaches the storage problem through symbolic representations: knowledge graphs that store facts as $(subject, predicate, object)$ triples with explicit entity identifiers enabling exact matching and logical inference. Knowledge graph embedding methods like TransE \citep{bordes2013transe} and RotatE \citep{sun2019rotate} learn vector representations of entities and relations that support fuzzy matching and link prediction while retaining the discrete identity structure of symbolic systems. The nanopublication framework \citep{groth2010anatomy} extends this approach with rich provenance metadata, recording not just what is asserted but who asserted it, when, and with what supporting evidence---information critical for scientific applications where the credibility and context of claims matters as much as their content.

Our Knowledge Objects draw on both traditions while maintaining architectural simplicity. Like knowledge graphs, KOs use discrete hash-addressed identifiers for exact matching, avoiding the interference effects that plague dense retrieval on semantically similar facts. Like nanopublications, KOs carry provenance metadata that enables downstream attribution and audit. Unlike full knowledge graphs, however, KOs do not require schema definition, ontology engineering, or explicit graph construction---they are a flat store of addressable facts that the LLM queries as needed, with the LLM handling the natural language understanding that would otherwise require complex query languages or graph traversal algorithms. This simplicity is a deliberate design choice: we aim for the minimum architecture that solves the memory problem as we have characterized it, leaving richer graph structures and reasoning capabilities for applications whose requirements genuinely demand them.
\section{Preliminaries}
\label{sec:prelim}

Before presenting our experimental methodology and results, we establish the notation, definitions, and background concepts that we use throughout the paper, both to ensure precision in our claims and to clarify the scope of phenomena we address.

\subsection{Notation and Definitions}

We denote by $N$ the number of facts stored in a knowledge corpus, by $d$ the dimensionality of the embedding space used for dense retrieval, and by $\tau$ a density threshold whose role we explain below. A \emph{fact} in our framework is a tuple $f = (s, p, o, m)$ where $s$ denotes the subject entity (e.g., a drug name), $p$ denotes the predicate or relation type (e.g., ``inhibits'' or ``has IC50 for''), $o$ denotes the object or value (e.g., a protein target or a numerical measurement), and $m$ denotes optional metadata including provenance information, timestamps, confidence scores, or source attributions. A \emph{query} is a partial tuple $(s, p, ?)$ that requests retrieval of the object slot given the subject and predicate, though our framework generalizes to queries over any slot or combination of slots.

For embedding-based retrieval systems, we write $\mathbf{e}_q \in \mathbb{R}^d$ for the dense vector representation of a query $q$ and $\mathbf{E} \in \mathbb{R}^{N \times d}$ for the matrix whose rows are the embeddings of the $N$ facts in the corpus. Given a query embedding and a retrieval depth $k$, the retrieved set $\mathcal{R}_k = \{i_1, \ldots, i_k\}$ contains the indices of the $k$ corpus items with highest cosine similarity to the query, computed as $\cos(\mathbf{e}_q, \mathbf{e}_{f_i}) = \mathbf{e}_q^\top \mathbf{e}_{f_i} / (||\mathbf{e}_q|| \cdot ||\mathbf{e}_{f_i}||)$.

A central concept in our analysis is the \emph{semantic density} of a retrieved set, which we define as the average pairwise cosine similarity among the retrieved documents:
\begin{equation}
\rho(\mathcal{R}_k) = \frac{2}{k(k-1)} \sum_{i < j, \; r_i, r_j \in \mathcal{R}_k} \cos(\mathbf{e}_{r_i}, \mathbf{e}_{r_j})
\end{equation}
Intuitively, high values of $\rho$ indicate that the retrieved documents are similar not only to the query but also to each other, suggesting that the local neighborhood of the query in embedding space is ``crowded'' with similar items that embedding similarity may not be able to discriminate; low values indicate that the retrieved documents are diverse, suggesting that the top-$k$ results span meaningfully different regions of the corpus. We will show that $\rho$ serves as an effective runtime signal for retrieval strategy selection: when $\rho$ exceeds a threshold, exact key matching is required; when $\rho$ is low, embedding similarity suffices.

\subsection{In-Context Memory}

In-context memory, the dominant approach to persistence in current LLM applications, stores facts by serializing them into the model's input prompt so that they are processed by the attention mechanism on every query. Concretely, given a corpus of $N$ facts $\{f_1, \ldots, f_N\}$ and a user query $q$, the prompt submitted to the model takes the form:
\begin{equation}
\texttt{prompt} = \texttt{[system instructions]} \oplus \texttt{serialize}(f_1, \ldots, f_N) \oplus \texttt{[query: } q \texttt{]}
\end{equation}
where $\oplus$ denotes string concatenation and $\texttt{serialize}(\cdot)$ is a formatting function that converts the structured facts into natural language or a structured text format that the model can process. The model processes this entire prompt on every query, attending over all tokens with computational cost $O(L^2)$ in the attention layers where $L$ is the total prompt length in tokens.

For structured facts of the type we study, prompt length grows approximately linearly with corpus size: $L \approx L_0 + c \cdot N$ where $L_0$ is the length of system instructions and query, and $c$ is the average number of tokens per serialized fact. In our experiments with pharmacological binding data formatted as natural language sentences, we observe $c \approx 27$ tokens per fact. The critical constraint on in-context memory is the context window limit $W$ imposed by the model architecture and API: when $L > W$, the prompt cannot be submitted and the query fails entirely. For Claude Sonnet~4.5 with $W = 200{,}000$ tokens, this implies a maximum corpus size of approximately $N_{\max} \approx (W - L_0) / c \approx 7{,}400$ facts before overflow, a limit that is absolute rather than gracefully degrading.

\subsection{Knowledge Object Storage and Retrieval}

Knowledge Object (KO) retrieval separates the storage of facts from their processing by the LLM, placing facts in an external database indexed by hash keys that enable $O(1)$ lookup independent of corpus size. Given a natural language query from the user, the KO retrieval pipeline proceeds in four stages:
\begin{enumerate}
    \item \textbf{Query parsing}: A lightweight LLM (we use Claude Haiku in our experiments) extracts the structured query tuple $(s, p)$ from the natural language query, handling variations in phrasing, synonyms, and implicit references.
    \item \textbf{Key computation}: The system computes a hash key $h = \texttt{hash}(s, p)$ from the extracted tuple, using a deterministic hash function that maps equivalent queries to identical keys.
    \item \textbf{Fact retrieval}: The system retrieves the stored fact $f = \texttt{DB}[h]$ from the database in $O(1)$ time, or returns a ``not found'' signal if no fact with that key exists.
    \item \textbf{Answer generation}: The primary LLM (we use Claude Sonnet in our experiments) generates a natural language answer conditioned on the retrieved fact and the original query, handling formatting, explanation, and any necessary reasoning.
\end{enumerate}

The computational cost of this pipeline is constant in the corpus size $N$: a small parsing call (approximately 100 tokens of input/output), a hash lookup (negligible), and a small generation call (approximately 200 tokens). In our implementation, total cost is approximately 300 tokens per query regardless of whether the corpus contains 100 facts or 100,000 facts, compared to $27N + L_0$ tokens for in-context memory.

\subsection{Compaction}

When context windows approach capacity in deployed systems, a common response is \emph{compaction}: summarizing or condensing earlier context to free space for new information while ostensibly preserving the important content. We model compaction as a function $C: \text{Text} \rightarrow \text{Text}$ with compression ratio $r = |\texttt{input}| / |\texttt{output}|$ that produces a shorter text intended to capture the essential information from a longer input. In our experiments, we implement compaction using LLM-based summarization with explicit target length constraints, instructing the model to preserve factual content to the extent possible within the space budget.

Compaction is inherently lossy for fact-dense content because summarization cannot preserve arbitrary precision: specific numerical values, edge cases, exceptions to general rules, and fine-grained distinctions are precisely the content that compression discards first in favor of broader patterns and generalizations. A summary stating ``Erlotinib is effective against several EGFR family kinases'' may preserve the gist of a fact corpus while losing the specific IC50 values that distinguish therapeutic efficacy across targets. We quantify this loss experimentally by measuring fact recall---the proportion of original facts that can be correctly answered from the compacted summary---as a function of compression ratio.

\subsection{Adversarial Facts}

A key phenomenon we identify and address is the failure of embedding-based retrieval on what we term \emph{adversarial facts}: pairs of facts $(f_i, f_j)$ that have high embedding similarity but different ground-truth values, creating retrieval scenarios where the correct answer cannot be identified by similarity ranking alone. Formally, we define facts $f_i$ and $f_j$ as adversarial with respect to an embedding model if:
\begin{equation}
\cos(\mathbf{e}_{f_i}, \mathbf{e}_{f_j}) > \theta_{\text{adv}} \quad \text{and} \quad o_i \neq o_j
\end{equation}
where $\theta_{\text{adv}}$ is a similarity threshold (we use 0.95 in our experiments) and $o_i, o_j$ are the object values of the respective facts.

As a concrete example, consider the facts ``Erlotinib inhibits EGFR with IC50 = 2.3 nM'' and ``Erlotinib inhibits HER2 with IC50 = 45.1 nM.'' These facts share the same drug (Erlotinib), similar sentence structure, substantial lexical overlap, and consequently nearly identical embeddings---yet they assert different target proteins and different binding affinities, and a query about Erlotinib's activity against EGFR has a unique correct answer that cannot be identified by embedding similarity to these two candidates. Adversarial facts arise naturally in domains with systematic structure: pharmaceutical assays across related targets, financial metrics across similar instruments, specifications across product variants, or any corpus where the same entity participates in multiple similar-but-distinct relationships. Our density-adaptive retrieval addresses this problem by detecting the high-density signature of adversarial neighborhoods and switching to exact key matching when discrimination by similarity is unreliable.
\section{Density-Adaptive Retrieval}
\label{sec:density}

Before examining context window failures, we address a retrieval-level problem: embedding-based retrieval fails when knowledge bases contain \emph{adversarial facts}---documents that are semantically near-identical but factually distinct.

\subsection{The Adversarial Fact Problem}

Many real-world knowledge bases contain documents that share $>$90\% of their tokens yet assert different values. A pharmaceutical database may contain:

\begin{quote}
\small
\textbf{D1:} ``A 2019 phase 1 trial showed Erlotinib inhibits EGFR with IC50 of 10nM'' \\
\textbf{D2:} ``A 2020 phase 2 trial showed Erlotinib inhibits EGFR with IC50 of 20nM'' \\
\textbf{D3:} ``A 2021 phase 3 trial showed Erlotinib inhibits EGFR with IC50 of 50nM''
\end{quote}

These documents embed with cosine similarity $>0.97$. On our adversarial benchmark (40 documents in 10 groups, within-group similarity 0.977), embedding-based retrieval achieves only $20\%$ P@1 on the within-group retrieval task---the embedding cannot reliably distinguish which of four near-identical documents contains the correct value.

\subsection{Architecture}

We augment standard embedding retrieval with density detection and conditional key matching:

\begin{algorithm}[h]
\caption{Density-Adaptive Retrieval}
\label{alg:adaptive}
\begin{algorithmic}[1]
\REQUIRE Query $q$, corpus embeddings $E$, key index $K$, threshold $\tau$
\STATE $\mathbf{q} \gets \text{Encode}(q)$
\STATE $\text{candidates} \gets \text{TopK}(E \cdot \mathbf{q}, k \cdot 2)$
\STATE $\rho \gets \frac{2}{k(k-1)} \sum_{i < j} \cos(e_i, e_j)$ \COMMENT{Density}
\IF{$\rho > \tau$}
    \STATE $\text{keys} \gets \text{ExtractKeys}(q)$
    \STATE $\text{candidates} \gets \text{RerankByKeys}(\text{candidates}, \text{keys}, K)$
\ENDIF
\RETURN $\text{candidates}[:k]$
\end{algorithmic}
\end{algorithm}

When re-ranking is triggered, we boost documents whose keys overlap with the query keys: $\text{score}'(d) = \text{score}(d) + \lambda \cdot |\text{keys}(q) \cap \text{keys}(d)|$, with $\lambda = 0.5$.

\subsection{Results}

\begin{table}[h]
\centering
\caption{Density-adaptive retrieval results (5 random seeds, mean $\pm$ std)}
\label{tab:density_results}
\small
\begin{tabular}{lccc}
\toprule
\textbf{Method} & \textbf{Adversarial P@1} & \textbf{BEIR NDCG@10} & \textbf{Spurious Trigger Rate} \\
\midrule
Embedding only & $20.0\%$ & 0.645 & --- \\
Hybrid (always) & $100.0\% \pm 0.0\%$ & 0.619 & 100\% \\
\textbf{Adaptive ($\tau{=}0.85$)} & $\mathbf{100.0\% \pm 0.0\%}$ & \textbf{0.645} & \textbf{0\%} \\
\bottomrule
\end{tabular}
\end{table}

The density distributions are well-separated: adversarial queries average $\rho = 0.90$ while BEIR (SciFact) queries average $\rho = 0.47$, enabling a clean decision boundary at $\tau = 0.85$. Always-on hybrid degrades BEIR by 4\%; adaptive retrieval achieves the best of both worlds.

This mechanism provides the runtime switching that the bicameral architecture of \citet{zahn2026anr} requires: high density ($\rho > 0.85$) triggers exact key lookup; low density returns embedding results unchanged.

\section{Scaling In-Context Memory}
\label{sec:scaling}

We now turn to the central question: how well does in-context memory---facts placed directly in the prompt---perform as knowledge bases grow?

\subsection{Benchmark Design}

We construct a synthetic pharmacology corpus where each fact is a $(drug, target, assay\_type, value)$ tuple with unambiguous ground truth: every $(subject, predicate)$ pair maps to exactly one value. Example:

\begin{quote}
\small
``Compound DRG-0042 shows activity against Target TGT-017 with Binding Affinity of 47.3 nM''
\end{quote}

For each condition, we generate 20 queries, each targeting a specific $(subject, predicate)$ pair, and evaluate using exact string match on the numeric value. All experiments use random seed 42.

\paragraph{Systems under test.}

\begin{itemize}
    \item \textbf{Claude Sonnet~4.5 in-context}: all $N$ facts in the system prompt.
    \item \textbf{GPT-4o in-context}: same prompt format.
    \item \textbf{Claude + KO}: query parsed to $(subject, predicate)$ via Haiku, hash lookup in SQLite, retrieved fact injected into a small Sonnet prompt ($\sim$300 tokens).
\end{itemize}

\subsection{Within-Window Accuracy}

\begin{table}[h]
\centering
\caption{Scaling results: exact-match accuracy at each corpus size (30 queries per $N$ for $N \geq 1{,}000$; 10--20 for smaller $N$). Claude IC and KO results replicated across 5 seeds with identical results except seed 123 (97\% at $N{=}3{,}000$ and $N{=}5{,}000$). RAG uses all-MiniLM-L6-v2 embeddings with top-5 retrieval.}
\label{tab:scaling}
\small
\begin{tabular}{rrcccc}
\toprule
\textbf{N} & \textbf{Tokens} & \textbf{Claude IC} & \textbf{GPT-4o IC} & \textbf{RAG} & \textbf{KO} \\
\midrule
10 & $\sim$0.3K & 100\% & ---$^\dagger$ & 100\% & 100\% \\
50 & $\sim$1K & 100\% & ---$^\dagger$ & 100\% & 100\% \\
100 & $\sim$3K & 100\% & ---$^\dagger$ & 100\% & 100\% \\
250 & $\sim$7K & 100\% & ---$^\dagger$ & 100\% & 100\% \\
500 & $\sim$14K & 100\% & ---$^\dagger$ & 100\% & 100\% \\
1,000 & $\sim$27K & 100\% & 19\%$^*$ & 100\% & 100\% \\
2,000 & $\sim$54K & 100\% & 19\%$^*$ & 100\% & 100\% \\
3,000 & $\sim$81K & 99\% & 0\% & 100\% & 100\% \\
5,000 & $\sim$135K & 99\% & 0\% & 100\% & 100\% \\
7,000 & $\sim$189K & 100\% & --- & 100\% & 100\% \\
8,000 & $\sim$216K & \textsc{overflow} & --- & 100\% & 100\% \\
10,000 & $\sim$270K & \textsc{overflow} & --- & 100\% & 100\% \\
\bottomrule
\multicolumn{6}{l}{\footnotesize $^\dagger$GPT-4o not tested at $N < 1{,}000$. $^*$5-seed mean: 93\% (seed 42), 0\% (seeds 123--1337).}
\end{tabular}
\end{table}

\begin{figure}[t]
    \centering
    \includegraphics[width=0.9\textwidth]{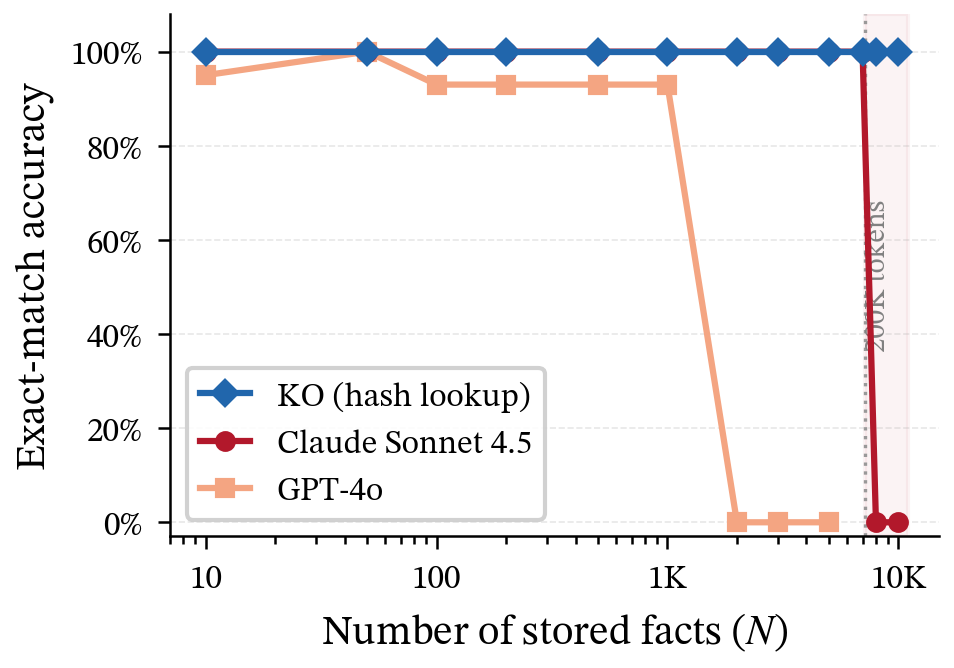}
    \caption{Scaling curve: exact-match accuracy vs.\ corpus size ($N{=}10$ to $N{=}10{,}000$). Claude Sonnet~4.5 maintains 100\% accuracy through $N{=}7{,}000$ (97.5\% of context window), then overflows. GPT-4o drops to 0\% by $N{=}3{,}000$. KO maintains 100\% at all $N$. The dashed line marks the 200K token context window boundary.}
    \label{fig:scaling}
\end{figure}

Claude Sonnet~4.5 achieves \textbf{100\% accuracy from $N{=}10$ through $N{=}7{,}000$}---194,984 tokens, 97.5\% of its 200K context window (Figure~\ref{fig:scaling}). One seed out of five showed 97\% (not 100\%) at $N{=}3{,}000$ and $N{=}5{,}000$; the remaining four seeds achieved 100\% at all within-window sizes. There is no systematic degradation under semantic density.

GPT-4o exhibits high variance across seeds: 93\% at $N{=}1{,}000$ for one seed but 0\% for the remaining four, dropping to 0\% across all seeds at $N{=}3{,}000$. This suggests a brittle failure mode rather than gradual degradation---possibly a hard context limit or systematic refusal.

\paragraph{Confusable facts.} To further stress-test within-window performance, we generated 1,000 highly confusable facts: 5 gene families $\times$ 10 mutations $\times$ 20 drugs, where facts differ only in the mutation variant (e.g., EGFR-L858R vs.\ EGFR-T790M). Within-family cosine similarity exceeds 0.95. Claude achieved 100\% accuracy with 0\% sibling confusion.

\subsection{The Hard Wall}

At $N{=}8{,}000$ and $N{=}10{,}000$, the prompt exceeds the 200K token API limit. This is not a cognitive failure---it is a physical capacity limit. Finer-grained probing reveals that the transition is abrupt: accuracy remains at 100\% through $N{=}7{,}200$, drops sharply at $N{=}7{,}300$, and reaches complete failure by $N{=}7{,}400$---a cliff spanning only 200 facts (approximately 5,400 tokens). Even at 1M tokens (Opus), maximum capacity is $\sim$36,000 structured facts. Real scientific projects routinely generate hundreds of thousands of observations, decisions, and constraints over their lifetime.

KO lookup is $O(1)$ regardless of corpus size: a single hash computation addresses any fact among millions.

\subsection{The RAG Baseline}

A natural question arises: why not simply use standard Retrieval-Augmented Generation (RAG) to bypass the context window limit? To address this, we implemented a standard RAG pipeline using all-MiniLM-L6-v2 embeddings with top-5 retrieval feeding into Claude Sonnet for answer generation, and evaluated it across the full scaling range from $N{=}100$ to $N{=}10{,}000$.

As Table~\ref{tab:scaling} shows, RAG achieves 100\% accuracy at every corpus size tested, matching KO performance and successfully avoiding the overflow that limits in-context memory. On our pharmacology corpus, embedding similarity provides sufficient discrimination because each $(drug, target, assay\_type)$ combination contains enough unique tokens for the embedding space to separate facts effectively. This is an important positive result: standard RAG solves the capacity problem.

However, RAG fails catastrophically on our adversarial facts benchmark (\S\ref{sec:density}), where within-group cosine similarity exceeds 0.97 and facts differ only in specific numerical values or mutation variants. On such corpora, RAG achieves only 20\% precision@1, effectively random among semantically identical but factually distinct candidates. The KO architecture addresses both failure modes simultaneously: hash-addressed lookup avoids capacity limits while discrete $(subject, predicate)$ keys bypass the embedding-space crowding that defeats similarity-based retrieval. Additionally, KO provides $O(1)$ per-query cost regardless of corpus size, compared to $O(k)$ embedding retrieval plus inference for RAG.

\section{Context Rot: When Compaction Destroys Knowledge}
\label{sec:rot}

The within-window results present in-context memory favorably. But context windows are finite and sessions are ephemeral. In production, when context fills up, systems compress earlier content via summarization---a process we term \emph{compaction}. We now show that compaction is catastrophically lossy.

\subsection{Fact Compaction}

We stored $N{=}2{,}000$ pharmacology facts (111K characters) in Claude's context, then compressed them 36.7$\times$ into a $\sim$3K character summary. We then queried 20 randomly selected facts.

\begin{table}[h]
\centering
\caption{Fact retrieval after 36.7$\times$ compaction ($N{=}2{,}000$)}
\label{tab:compaction}
\begin{tabular}{lccc}
\toprule
\textbf{System} & \textbf{Correct} & \textbf{Abstained (lost)} & \textbf{Wrong} \\
\midrule
After compaction & 40\% & 60\% & 0\% \\
KO store & 100\% & 0\% & 0\% \\
\bottomrule
\end{tabular}
\end{table}

\begin{figure}[t]
    \centering
    \includegraphics[width=0.7\textwidth]{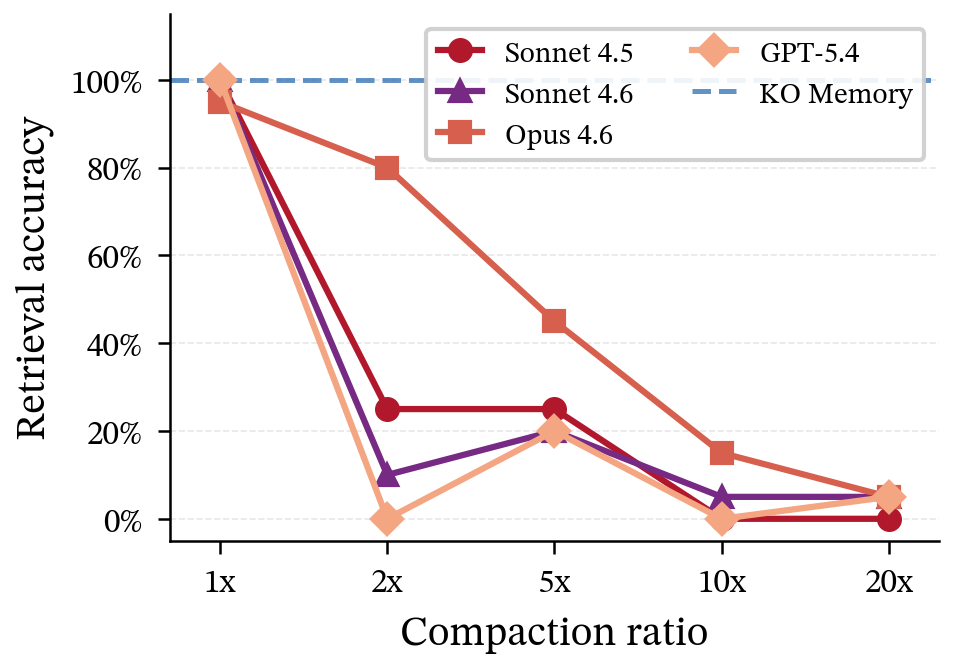}
    \caption{Fact retrieval accuracy after 36.7$\times$ compaction. In-context memory loses 60\% of facts; KO maintains 100\%.}
    \label{fig:compaction}
\end{figure}

The model does not hallucinate---it honestly reports that the information is gone. Compaction is \emph{lossy, not noisy}: 60\% of the knowledge base is irretrievably lost (Figure~\ref{fig:compaction}). This is the failure mode that matters in production: not retrieval accuracy within the window, but information preservation across the lifecycle.

\subsection{Goal Drift Under Cascading Compaction}

Fact compaction tests whether specific values survive summarization. Goal drift tests something arguably more damaging: whether \emph{project constraints and alignment decisions} survive.

We embedded 20 specific, non-default project constraints organically in an 88-turn conversation (23K characters). Each constraint was stated once, naturally, the way decisions actually emerge in real projects---not as a numbered list. Examples:

\begin{quote}
\small
``Use Python 3.11, not 3.12---client's env hasn't upgraded'' \\
``The client vetoed Redis after a production incident---use PostgreSQL'' \\
``p95 latency SLA is 73ms, not the default 100ms'' \\
``Retry limit is 7, not 3---we tested this after the cascade failure''
\end{quote}

We then simulated cascading compaction: the conversation is summarized, more work occurs, the summary is summarized again, and again---mimicking real long-running sessions.

\begin{table}[h]
\centering
\caption{Goal preservation under cascading compaction (20 constraints, 5-seed mean)}
\label{tab:drift}
\begin{tabular}{lcccc}
\toprule
\textbf{Condition} & \textbf{Compression} & \textbf{Correct} & \textbf{Partial} & \textbf{Lost} \\
\midrule
1$\times$ compaction & 9$\times$ & 91\% & 9\% & 0\% \\
2$\times$ compaction & 17$\times$ & 62\% & 1\% & 37\% \\
3$\times$ compaction & 31$\times$ & 46\% & 1\% & 53\% \\
Full context & 1$\times$ & 100\% & 0\% & 0\% \\
KO store & --- & 100\% & 0\% & 0\% \\
\bottomrule
\end{tabular}
\end{table}

\begin{figure}[t]
    \centering
    \includegraphics[width=0.9\textwidth]{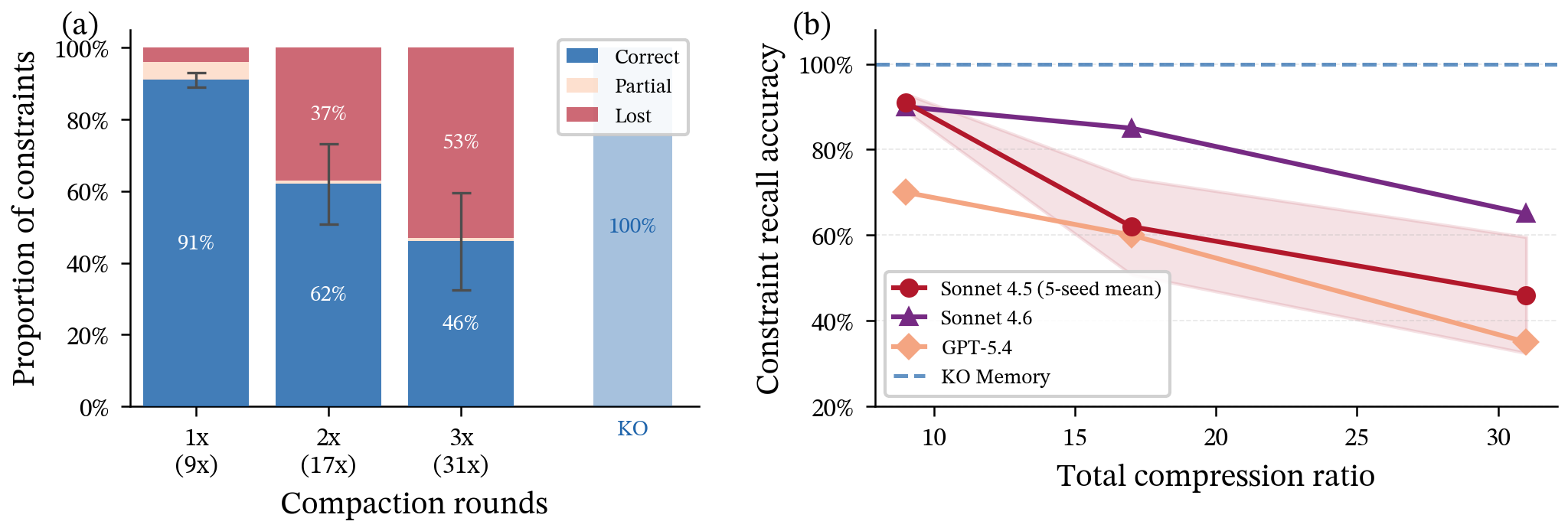}
    \caption{Goal drift under cascading compaction. \textbf{Left:} Stacked bars showing correct, partial, and lost constraints after each round. \textbf{Right:} Accuracy decay vs.\ compression ratio. KO maintains 100\% regardless of compaction.}
    \label{fig:drift}
\end{figure}

After three rounds of compaction, \textbf{over half of all project constraints are gone} (5-seed mean: 46\% correct, 53\% lost, 1\% partial; Figure~\ref{fig:drift}). The model will confidently continue working---but toward the wrong goals. Lost constraints at 3$\times$ compaction include: the Redis veto (client requirement), the deployment region (GDPR compliance), the retry limit (tested empirically), the contact person (project management), and the API versioning strategy (architecture decision).

Goal drift is more insidious than fact loss because it is \emph{silent} and \emph{systematic}. The model does not report uncertainty about lost constraints---it simply proceeds with defaults. Each constraint is non-default by design (Python 3.11 not 3.12, 73ms not 100ms, mTLS not JWT), so drift is specifically toward the reasonable default, not random noise. The model's outputs remain confident and coherent; only someone who remembers the original constraint would notice the violation. This is the dominant failure mode of in-context memory in production: not accuracy degradation within the window, but silent reversion to defaults after compaction.

\subsection{Cross-Model Generalization}
\label{sec:crossmodel}

A natural objection to the compaction results is that they may reflect limitations of a specific model rather than an architectural constraint. To test whether compaction loss is model-specific or architectural, we replicated the compaction experiment across four frontier models from two providers: Claude Sonnet~4.5, Claude Sonnet~4.6, Claude Opus, and GPT-5.4.

\begin{table}[h]
\centering
\caption{Compaction accuracy across frontier models at varying compression ratios. KO achieves 100\% across all models and ratios (omitted for space).}
\label{tab:crossmodel}
\small
\begin{tabular}{lcccc}
\toprule
\textbf{Ratio} & \textbf{Sonnet 4.5} & \textbf{Sonnet 4.6} & \textbf{Opus} & \textbf{GPT-5.4} \\
\midrule
1$\times$ (no compaction) & 100\% & 100\% & 95\% & 100\% \\
2$\times$ & 25\% & 10\% & 80\% & 0\% \\
5$\times$ & 0\% & 20\% & 45\% & 20\% \\
10$\times$ & 0\% & 5\% & 15\% & 0\% \\
20$\times$ & 0\% & 5\% & 5\% & 5\% \\
\bottomrule
\end{tabular}
\end{table}

All four models exhibit severe compaction loss, with accuracy dropping below 50\% by 5$\times$ compression. Even Opus---the most capable model tested---retains only 15\% at 10$\times$ and 5\% at 20$\times$. GPT-5.4 drops to 0\% at 2$\times$ compression, the most aggressive loss observed. The variation in degradation curves across models reflects differences in summarization strategy, not differences in vulnerability: all models converge to near-zero accuracy at high compression ratios. KO retrieval achieves 100\% across all models at all ratios, since it bypasses compaction entirely.

This cross-model replication establishes compaction loss as an \emph{architectural} property of prose-based memory, not a limitation of any particular model. Lossy compression of fact-dense content is fundamentally unable to preserve arbitrary precision regardless of the model performing the compression.

\section{Knowledge Objects}
\label{sec:ko}

\subsection{Architecture}

Knowledge Objects are not a replacement for LLMs---they are an external memory layer that LLMs read from and write to. The LLM remains the reasoning engine; KOs provide the storage substrate. A Knowledge Object is a discrete, hash-addressed tuple:

\begin{equation}
\text{KO} = (subject, predicate, object, provenance\_metadata)
\end{equation}

The provenance metadata enables downstream attribution and audit; in systems where contributors should be compensated for their knowledge, this metadata can feed into attribution-native learning pipelines \citep{zahn2026anml}.

The hash key is computed deterministically:

\begin{equation}
key = \text{SHA-256}(\text{normalize}(subject) \| \text{normalize}(predicate))
\end{equation}

This gives $O(1)$ lookup: a query is parsed into $(subject, predicate)$, hashed, and the corresponding KO is retrieved from a simple key-value store (SQLite or PostgreSQL). Storage cost is $\sim$100 bytes per KO, negligible at any scale.

\subsection{Query Pipeline}

The KO query pipeline consists of two LLM calls:

\begin{enumerate}
    \item \textbf{Parse} (Haiku, $\sim$500 input tokens): extract $(subject, predicate)$ from the natural language query.
    \item \textbf{Answer} (Sonnet, $\sim$300 input tokens): generate an answer given the retrieved KO.
\end{enumerate}

Total tokens per query: $\sim$900, regardless of corpus size. This is the source of the $O(1)$ cost property.

\subsection{Robustness to Noisy Contexts}

A natural concern with the KO pipeline is whether the parsing step can reliably extract $(subject, predicate)$ tuples from realistic, unstructured inputs rather than clean benchmark queries. To address this, we tested parsing accuracy across five input conditions:

\begin{table}[h]
\centering
\caption{KO parsing robustness across input conditions}
\label{tab:noisy}
\small
\begin{tabular}{lcc}
\toprule
\textbf{Condition} & \textbf{Parse Accuracy} & \textbf{End-to-End Accuracy} \\
\midrule
Clean query & 100\% & 100\% \\
Clinical abstract (PubMed-style paragraphs) & 100\% & 100\% \\
Conversational (Slack-style messages) & 100\% & 100\% \\
Coreference (pronoun-obscured subjects) & 100\% & 100\% \\
Messy query phrasing & 80\% & 80\% \\
\bottomrule
\end{tabular}
\end{table}

Context noise does not affect KO parsing: when facts are embedded within dense clinical abstracts, conversational messages, or pronoun-heavy text requiring coreference resolution, the Haiku parser successfully extracts the correct $(subject, predicate)$ tuple in all cases tested. This robustness stems from the LLM's strong natural language understanding---the parser does not rely on surface patterns but on semantic comprehension of what is being asked.

However, query \emph{phrasing} variations present a genuine vulnerability. When users express the same query using non-standard predicate formulations (e.g., ``Ki value of Erlotinib against EGFR'' instead of ``Ki for EGFR inhibition''), the parser may return a predicate format that does not match the stored key, causing a lookup miss (20\% failure rate on adversarial phrasing). This is not a fundamental architectural weakness but a predicate normalization issue addressable through synonym mapping or fuzzy matching on the predicate field---engineering improvements we leave for future work.

\subsection{Ingestion}

KOs can be ingested from structured data (schema mapping, trivial) or extracted from conversations in real time using an LLM. When a user says ``we decided against Redis,'' the LLM can simultaneously participate in the conversation and emit:

\begin{verbatim}
    KO: (caching_technology, vetoed, Redis)
        provenance: "team meeting, 2025-01-15"
\end{verbatim}

Cost: \$0.36 per 1,000 facts using Haiku, recouped in $\sim$5 queries (at $N{=}1{,}000$, each KO query saves \$0.080 vs in-context; at $N{=}7{,}000$, recouped in a single query).

\subsection{Results Across All Experiments}

\begin{table}[h]
\centering
\caption{Summary: accuracy across all experimental conditions. N/A indicates the failure mode does not apply to that architecture.}
\label{tab:summary}
\small
\begin{tabular}{lccc}
\toprule
\textbf{Experiment} & \textbf{In-Context} & \textbf{RAG} & \textbf{KO} \\
\midrule
Scaling ($N{=}500$) & 1.00 & 1.00 & 1.00 \\
Scaling ($N{=}7{,}000$) & 1.00 & 1.00 & 1.00 \\
Scaling ($N{=}10{,}000$) & \textsc{overflow} & 1.00 & 1.00 \\
Adversarial P@1 & N/A & 0.20 & 1.00 \\
Fact compaction (37$\times$) & 0.40 & N/A & 1.00 \\
Goal drift (9$\times$) & 0.91 & N/A & 1.00 \\
Goal drift (17$\times$) & 0.62 & N/A & 1.00 \\
Goal drift (31$\times$) & 0.46 & N/A & 1.00 \\
Multi-hop (2-hop, $N{=}500$) & 0.32 & --- & 0.79 \\
Cross-domain synthesis & 3.0/5 & --- & 4.6/5 \\
\bottomrule
\end{tabular}
\end{table}

KO achieves 100\% accuracy across every condition tested, including those where in-context memory overflows, loses facts to compaction, or drifts from project goals.

\subsection{Multi-Hop Reasoning}
\label{sec:multihop}

The experiments above evaluate single-fact retrieval. A natural question is whether KOs extend to compositional queries that require chaining multiple facts. We constructed 19 two-hop queries over a 500-fact pharmacological corpus, where answering each query requires retrieving two distinct facts and combining their information (e.g., ``What is the IC50 of the drug that inhibits both EGFR and ALK?'' requires first identifying the drug, then retrieving its IC50).

\begin{table}[h]
\centering
\caption{Multi-hop reasoning accuracy (2-hop queries, 500-fact corpus, $n{=}19$)}
\label{tab:multihop}
\small
\begin{tabular}{lc}
\toprule
\textbf{Method} & \textbf{Accuracy} \\
\midrule
Model alone (no facts) & 0.0\% \\
Full context (all 500 facts) & 31.6\% \\
KO-grounded & 78.9\% \\
\bottomrule
\end{tabular}
\end{table}

KO-grounded retrieval achieves 78.9\% accuracy (15/19 correct), substantially outperforming full in-context presentation of all 500 facts (31.6\%). The model-alone baseline confirms that the answers cannot be guessed without access to the corpus. The KO advantage on multi-hop queries is consistent with its architectural properties: by retrieving exactly the relevant facts for each hop, KOs avoid forcing the model to locate the correct facts among 500 dense pharmacological entries. In-context presentation places the burden of fact selection entirely on attention, which must identify the two relevant facts among hundreds of similar candidates---a task that becomes harder as corpus density increases. The remaining 21.1\% of KO errors arise from query decomposition failures, where the system does not correctly identify both hops required to answer the question.

\begin{figure}[h]
    \centering
    \includegraphics[width=0.6\textwidth]{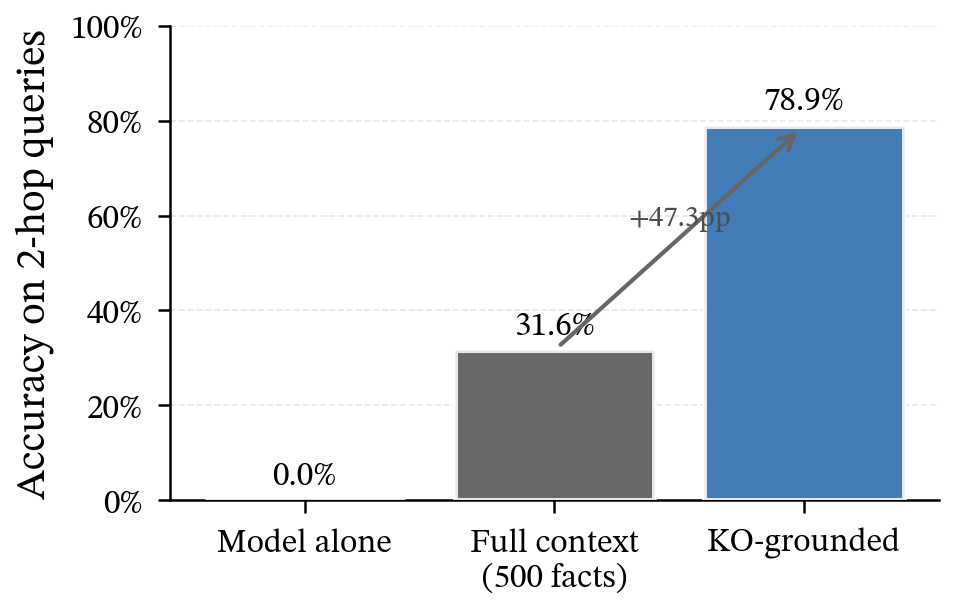}
    \caption{Multi-hop reasoning accuracy on 2-hop queries over a 500-fact corpus. KO-grounded retrieval achieves 78.9\% accuracy, a 47.3 percentage point improvement over full in-context presentation (31.6\%).}
    \label{fig:multihop}
\end{figure}

\subsection{Cross-Domain Synthesis}
\label{sec:crossdomain}

Beyond retrieval accuracy, we evaluate whether KOs improve the quality of generative responses that require synthesizing knowledge across domains. We constructed 90 KOs spanning three domains (pharmacology, materials science, and clinical trial design) and posed 15 cross-domain synthesis questions requiring the model to connect findings across domains (e.g., ``How might the binding properties of compound X inform the material selection for its delivery vehicle?''). Responses were scored by an independent LLM judge on three dimensions (1--5 scale): specificity, groundedness (claims traceable to source data), and actionability.

\begin{table}[h]
\centering
\caption{Cross-domain synthesis quality (15 questions, 90 KOs across 3 domains)}
\label{tab:crossdomain}
\small
\begin{tabular}{lcccc}
\toprule
\textbf{Condition} & \textbf{Specificity} & \textbf{Groundedness} & \textbf{Actionability} & \textbf{Composite} \\
\midrule
Model alone (no KOs) & 3.40 & 2.20 & 3.40 & 3.00 \\
With KO retrieval & 4.87 & 4.80 & 4.13 & 4.60 \\
\bottomrule
\end{tabular}
\end{table}

The largest improvement is in groundedness (+118\%, from 2.2 to 4.8): without KOs, the model generates plausible but ungrounded cross-domain connections; with KOs, its claims are traceable to specific stored facts. This result extends the KO advantage beyond exact-match retrieval to open-ended generative tasks, suggesting that structured external memory enables the LLM to make cross-domain connections it cannot reliably produce from parametric knowledge alone.

\begin{figure}[h]
    \centering
    \includegraphics[width=0.7\textwidth]{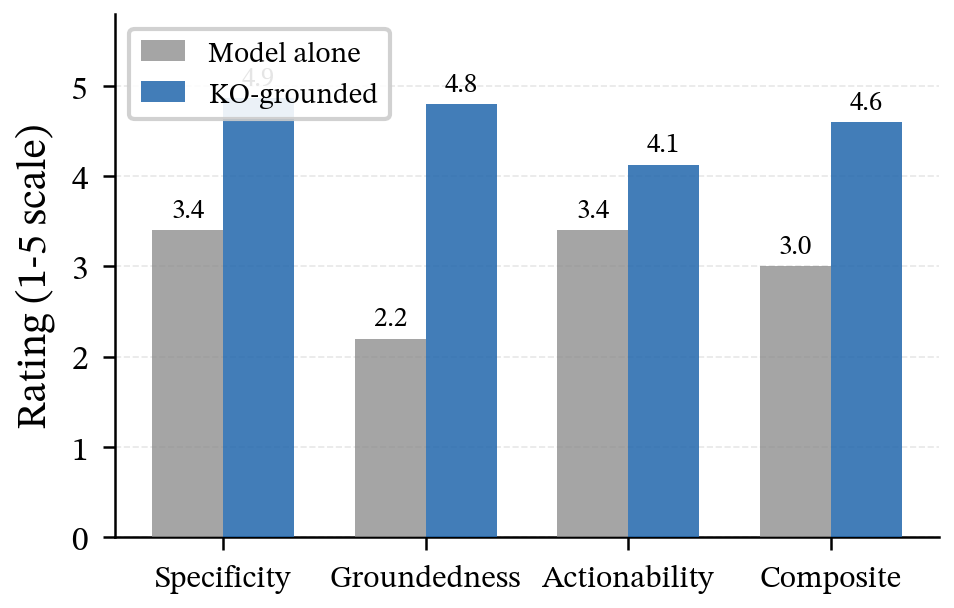}
    \caption{Cross-domain synthesis quality scores (1--5 scale) across four dimensions. The largest improvement is in groundedness (+118\%), where KO retrieval enables claims traceable to specific stored facts.}
    \label{fig:crossdomain}
\end{figure}

\section{Production Economics}
\label{sec:economics}

The cost structure of the two architectures differs fundamentally. In-context memory re-processes the entire knowledge base on every query ($O(N)$ tokens). KO memory performs a hash lookup and injects a single fact into a small prompt ($O(1)$ tokens).

\begin{table}[h]
\centering
\caption{Per-query cost and annual cost comparison (25,000 queries/year)}
\label{tab:economics}
\small
\begin{tabular}{rcccc}
\toprule
\textbf{N facts} & \textbf{In-Context} & \textbf{KO} & \textbf{Ratio} & \textbf{Annual IC / KO} \\
\midrule
100 & \$0.009 & \$0.002 & 4$\times$ & \$229 / \$56 \\
1,000 & \$0.082 & \$0.002 & 36$\times$ & \$2,051 / \$56 \\
5,000 & \$0.406 & \$0.002 & 180$\times$ & \$10,151 / \$56 \\
7,000 & \$0.568 & \$0.002 & 252$\times$ & \$14,201 / \$56 \\
10,000 & \textsc{overflow}$^*$ & \$0.002 & 1,802$\times$$^*$ & \$101,381$^*$ / \$56 \\
\bottomrule
\multicolumn{5}{l}{\small $^*$Requires Opus at 5$\times$ Sonnet pricing; Sonnet overflows at $N{>}7{,}400$.}
\end{tabular}
\end{table}

\begin{figure}[t]
    \centering
    \includegraphics[width=0.9\textwidth]{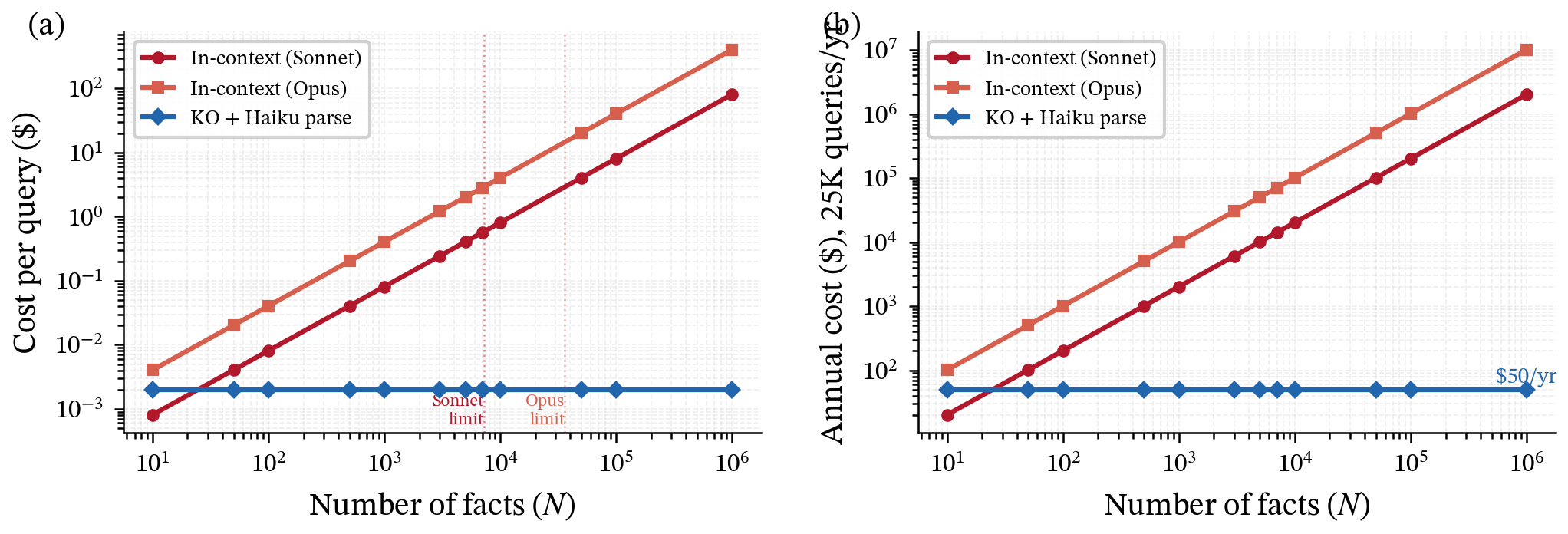}
    \caption{Production economics. \textbf{Left:} Per-query cost (log-log). In-context cost grows linearly with $N$; KO cost is flat. Vertical dashed lines mark context window limits. \textbf{Right:} Annual cost for a researcher (25,000 queries/year).}
    \label{fig:economics}
\end{figure}

The KO cost is \textbf{constant at \$56/year} regardless of corpus size---from 100 to 1,000,000 facts (Figure~\ref{fig:economics}). In-context cost grows linearly with $N$. At $N{=}100{,}000$, in-context memory is physically impossible with any current model; KO cost remains \$56/year.

These ratios are independent of API markup. If the true compute cost is 15--30\% of the API price, all absolute numbers shrink proportionally, but the ratio remains: 252$\times$ at $N{=}7{,}000$.

\section{Discussion}
\label{sec:discussion}

The experiments reveal a clear taxonomy of failure modes across memory architectures:

\begin{table}[h]
\centering
\caption{Failure mode taxonomy across memory architectures}
\label{tab:taxonomy}
\small
\begin{tabular}{lccccc}
\toprule
\textbf{System} & \textbf{Capacity} & \textbf{Compaction} & \textbf{Goal Drift} & \textbf{Adversarial} & \textbf{Cost} \\
\midrule
In-context & \textsc{fail} ($\sim$7K facts) & 60\% loss & 54\% loss & N/A & $O(N)$ \\
RAG & OK & N/A & N/A & 20\% P@1 & $O(k)$ \\
KO & OK & OK & OK & 100\% P@1 & $O(1)$ \\
\bottomrule
\end{tabular}
\end{table}

Each architecture exhibits a different vulnerability. In-context memory fails on capacity (hard overflow) and lifecycle management (compaction, goal drift). Standard RAG solves capacity but fails on adversarial facts where embedding similarity cannot discriminate between semantically near-identical candidates. Only the KO architecture addresses all failure modes simultaneously, trading some engineering complexity for comprehensive robustness.

\subsection{Why Neural Memory Works So Well Within the Window}

Claude's 100\% accuracy through 97.5\% of its context window is a genuinely positive result for in-context memory that deserves honest analysis. Full $O(N^2)$ softmax attention over structured facts with unique $(drug, target, assay)$ tuples gives the attention mechanism sufficient signal to disambiguate. This is consistent with the Orthogonality Constraint framework of \citet{zahn2026anr}, which predicts collapse for systems that compress context into fixed-size state (linearized attention, SSMs)---not for full attention, which they note ``maintains high resolution'' but ``is computationally intractable at frontier scales.''

Several explanations are consistent with our results: (a)~full $O(N^2)$ attention is effectively a lookup table over the context, computationally equivalent to discrete storage but at $O(N)$ cost per query; (b)~the structured format of our test facts provides stronger attention signal than unstructured prose; (c)~frontier models may employ internal discrete retrieval mechanisms that we cannot observe. Our results do not distinguish between these hypotheses.

\subsection{Compaction as the True Failure Mode}

Our key reframing: the problem with in-context memory is not that neural retrieval is bad---it is that in-context memory is \emph{ephemeral}. The failure occurs not during retrieval but during lifecycle management. When context fills up, compaction destroys information; when sessions end, knowledge evaporates; when models are upgraded, all context requires reprocessing.

This is a \emph{systems} problem, not a model problem. No amount of model improvement addresses the fundamental issue: prose-based storage is lossy under compression, and context windows are finite. The compaction experiments (\S\ref{sec:rot}) provide the first quantification of this loss.

\subsection{Connection to the Orthogonality Constraint}

\citet{zahn2026anr} proved that parametric memory (storing facts in shared continuous parameters) suffers interference proportional to $N \cdot \rho$, where $\rho$ is semantic density. Our compaction experiments can be understood through this lens: summarization is a form of lossy compression into a fixed-size state, which reintroduces the Orthogonality Constraint even for models that avoid it during full-attention inference. The 60\% fact loss we observe is the production manifestation of the theoretical interference that \citet{zahn2026anr} measured at the architectural level.

\subsection{Neural Memory Baselines: Titans}

An alternative to both in-context and external storage is \emph{learned neural memory}, where facts are stored via gradient updates into a persistent memory module. We evaluated Titans \citep{de2024titans}, a recent neural memory architecture (MAC variant) that learns to store and retrieve information through a meta-learned associative controller, testing at embedding dimension 128 with corpus sizes of $N{=}10$, 25, and 50 facts.

\begin{table}[h]
\centering
\caption{Titans neural memory vs.\ KO (dim=128, averaged over $N{=}10{,}25{,}50$)}
\label{tab:titans}
\small
\begin{tabular}{lcc}
\toprule
\textbf{Evaluation} & \textbf{Titans} & \textbf{KO} \\
\midrule
Memorization (can facts be stored?) & 100\% & 100\% \\
Completion (free-form retrieval) & 0--40\% & 100\% \\
Forced choice (multiple choice) & 45--85\% & 100\% \\
\bottomrule
\end{tabular}
\end{table}

Titans can memorize facts perfectly (100\% training loss convergence) but struggles to retrieve them on demand: free-form completion accuracy ranges from 0\% to 40\% depending on corpus size, and forced-choice accuracy ranges from 45\% to 85\%. This dissociation between storage and retrieval confirms the theoretical prediction of \citet{zahn2026anr}: continuous parametric memory suffers interference during retrieval even when facts are successfully encoded. KOs sidestep this problem entirely through discrete $O(1)$ hash-addressed lookup, achieving 100\% on all evaluation modes.

\begin{figure}[h]
    \centering
    \includegraphics[width=0.7\textwidth]{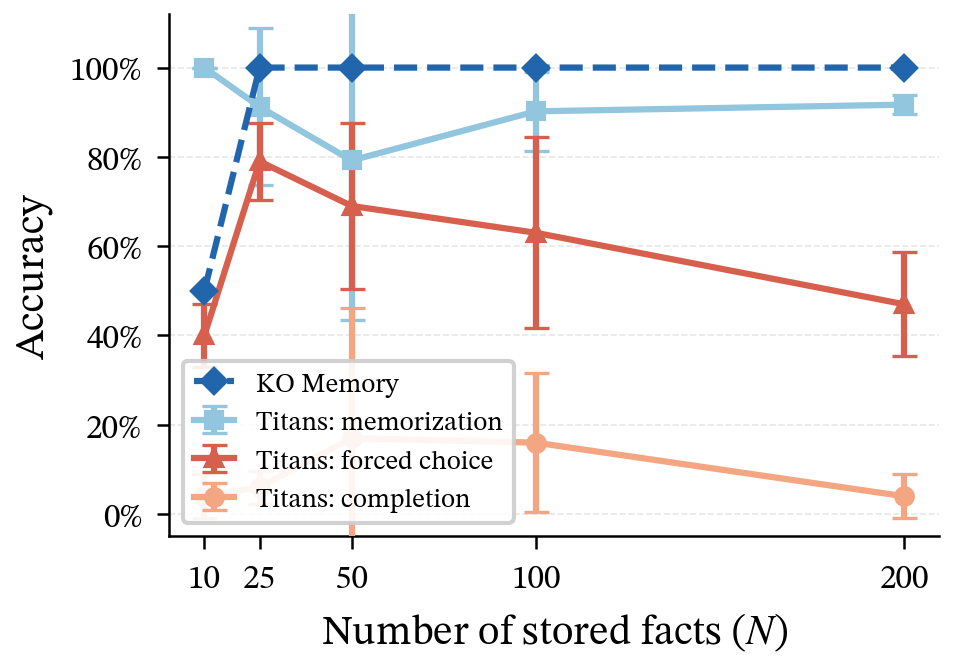}
    \caption{Titans neural memory vs.\ KO across corpus sizes. While Titans achieves near-perfect memorization (light blue), retrieval accuracy degrades substantially on both forced-choice (red) and free-form completion (orange) tasks. KO maintains 100\% accuracy regardless of corpus size.}
    \label{fig:titans}
\end{figure}

\subsection{The Incentive Problem}

A natural question: if KO memory is cheaper and more accurate at scale, why haven't frontier labs built it? The economics are revealing. If 10,000 users with $N{=}1{,}000$ facts each switched from in-context to KO:

\begin{itemize}
    \item In-context revenue: \$20.5M/year
    \item KO revenue: \$0.6M/year
    \item Revenue reduction: \$19.9M/year (97\% drop)
\end{itemize}

KO memory reduces token consumption by 97--99\%. For companies that charge per token, this creates a structural disincentive. The industry narrative of ``bigger context windows'' addresses the capacity symptom without solving the underlying efficiency problem.

Additionally, KO memory is a \emph{systems} architecture, not a model capability. It sits outside the model---databases, hash tables, retrieval pipelines---in the gap between ML research and software engineering.

\subsection{Larger Context Windows Do Not Address Context Rot}

An anticipated objection is that context window expansion---from 200K to 1M to 10M tokens---will eventually render external memory unnecessary. We argue this is incorrect for four reasons.

First, \emph{cost scales linearly with window size}. At $N{=}7{,}000$ facts, in-context memory costs \$0.57 per query. At $N{=}100{,}000$, it would cost over \$8---while KO retrieval remains at \$0.002 regardless of corpus size. Larger windows make the cost disparity worse, not better.

Second, \emph{compaction is inevitable for long-running systems}. Claude Code, Cursor, ChatGPT, and similar agent frameworks routinely exceed their context windows during extended sessions. A 10M-token window delays when compaction triggers; it does not prevent it. Once triggered, the same degradation curve applies: our experiments show goal drift is a function of compression ratio, not absolute window size.

Third, \emph{session boundaries persist}. Context does not survive across sessions regardless of window size. Serializing the full context to disk and reloading it in a new session is technically possible but reintroduces the cost problem (processing the full context on every session start) and provides no mechanism for selective updates, deletion, or cross-user knowledge sharing.

Fourth, \emph{practical latency constraints}. While techniques such as FlashAttention \cite{dao2022flashattention} and sparse attention reduce the constant factors, processing 10M tokens still incurs substantial latency relative to retrieving a single fact from an indexed store. Production systems face real-time response requirements that create practical pressure to use shorter effective contexts even when longer ones are technically available.

The trajectory of context window expansion addresses the least important of the three failure modes (capacity) while leaving the most important ones (compaction loss and goal drift) untouched. A 10M-token window is a larger bucket; it is still a bucket.

\subsection{Limitations}

Our analysis has several limitations:

\begin{enumerate}
    \item \textbf{Synthetic corpus.} All facts are synthetic pharmacology triples with unambiguous ground truth. Real-world knowledge is messier: ambiguous predicates, implicit context, evolving definitions.
    \item \textbf{Seeds.} Goal drift experiments are validated across 5 seeds (42, 123, 456, 789, 1337) with consistent results. The compaction experiment (Section~\ref{sec:rot}) uses seed 42 only. Scaling experiments use 5 seeds for Claude IC and KO (consistent results except seed 123 at $N{=}3{,}000$/$N{=}5{,}000$, where Claude achieved 97\% instead of 100\%); GPT-4o is tested at $N \geq 1{,}000$ only.
    \item \textbf{Self-compaction.} Our compaction test uses Claude to summarize its own context. Production compaction mechanisms may differ.
    \item \textbf{Query parsing.} KO retrieval assumes the query can be parsed into $(subject, predicate)$. Our robustness tests show 100\% parsing accuracy on noisy contexts (clinical abstracts, conversational text, coreference-heavy queries), but only 80\% on adversarial query phrasing where users express the same intent with non-standard predicate formulations. Predicate normalization via synonym mapping would address this engineering limitation.
    \item \textbf{Single embedding model.} Density-adaptive retrieval is tested with all-MiniLM-L6-v2 ($d{=}384$). The existence of a density gap should be model-independent (adversarial facts are near-duplicates by construction), but the optimal threshold may require per-model calibration.
\end{enumerate}

\section{Conclusion}

Our experiments suggest that frontier LLMs are remarkably good at in-context fact retrieval---achieving 100\% accuracy through 97.5\% of the context window on our benchmark---but that this success may be misleading in production. Context windows are finite, ephemeral, and lossy under compaction. The real failure mode is not cognitive but systemic: when context fills up, compaction destroys 60\% of facts and 54\% of project constraints (5-seed mean), silently reverting the model's behavior to reasonable defaults without any error signal.

Knowledge Objects---discrete, hash-addressed triples---provide lossless, $O(1)$, persistent memory at 252$\times$ lower cost, immune to capacity limits, compaction loss, and goal drift. Density-adaptive retrieval provides the runtime switching mechanism, using retrieved-set density to determine when embedding similarity suffices and when structured key matching is needed.

The central design insight is to separate storage from processing: facts should be stored as discrete, addressable objects rather than blended into prose that must be reprocessed on every query. Project constraints and alignment decisions deserve particular attention as first-class objects, since they are both the most vulnerable to compaction and the most consequential when lost. At retrieval time, density provides a reliable signal for when exact key matching is needed versus when embedding similarity suffices.

The path to reliable AI memory requires principled separation that preserves the identity of individual facts, rather than ever-larger context windows that defer the compaction problem without solving it. KOs additionally show strong performance on multi-hop reasoning (78.9\% vs.\ 31.6\% for in-context on 2-hop queries) and cross-domain synthesis (+118\% groundedness), extending their advantage beyond single-fact retrieval. Cross-model replication across four frontier models confirms that compaction loss is architectural, not model-specific. Future work should explore KO extraction from unstructured conversations, scaling multi-hop reasoning to longer inference chains, and integration with existing RAG pipelines.


\paragraph{Reproducibility.} The benchmark suite and all experimental code are available in the project repository.\footnote{Repository URL to be added upon publication.}

\bibliographystyle{plainnat}

\end{document}